\documentclass[10pt,twocolumn,letterpaper]{article}

\usepackage{cvpr}
\usepackage{times}
\usepackage{epsfig}
\usepackage{graphicx}
\usepackage{amsmath}
\usepackage{commath}
\usepackage{amssymb}
\usepackage{import}

\usepackage{epstopdf}
\usepackage{url}

\usepackage{times}

\usepackage{url}

\usepackage{amsfonts}

\usepackage{graphicx}

\usepackage{parskip}

\usepackage{multirow}

\usepackage{psfrag}

\usepackage{verbatim}

\usepackage{wrapfig}

\usepackage{paralist}

\usepackage{enumitem}
\setdescription{leftmargin=0.25cm}

\usepackage[usenames,dvipsnames]{color}

\newcounter{mycounter}

\usepackage[]{algorithm2e}
\makeatletter
\newcommand{\Removelatexerror}{\let\@latex@error\@gobble}
\makeatother
\newcommand{\myalgorithm}{%
\SetInd{0.5em}{0.5em}
\let\oldnl\nl
\newcommand{\nonl}{\renewcommand{\nl}{\let\nl\oldnl}}
\newcommand{\pushline}{\Indp}
\newcommand{\popline}{\Indm\dosemic}
\begingroup
\Removelatexerror 
\begin{algorithm*}[H]
\end{algorithm*}
\endgroup}

\usepackage{fixltx2e}

\makeatletter
\renewcommand{\paragraph}{%
  \@startsection{paragraph}{4}%
  {\z@}{0.25ex \@plus 1ex \@minus .2ex}{-1em}%
  {\normalfont\normalsize\bfseries}%
}
\makeatother

\usepackage[pagebackref=true,breaklinks=true,letterpaper=true,colorlinks,bookmarks=false]{hyperref}
\hypersetup{
	colorlinks=true,
	citecolor=blue,  
	urlcolor=red,
}
\cvprfinalcopy 

\def\cvprPaperID{561} 

\ifcvprfinal\fi
\begin{document}

\title{CSGNet: Neural Shape Parser for Constructive Solid Geometry}

\author{Gopal  Sharma \quad Rishabh Goyal \quad Difan Liu \quad Evangelos Kalogerakis \quad Subhransu Maji \\
University of Massachusetts, Amherst\\
{\tt\small \{gopalsharma,risgoyal,dliu,kalo,smaji\}@cs.umass.edu}
}

\maketitle

\begin{abstract}

We present a neural architecture that takes as input a 2D or 3D shape
and outputs a program that generates the shape. The instructions in our program are based on constructive solid geometry
principles, i.e., a set of boolean operations on shape primitives
defined recursively.
Bottom-up techniques for this shape parsing task rely on primitive detection and are inherently slow since the search space over possible primitive combinations is large.
In contrast, our model uses a recurrent neural network that parses the input shape 
in a top-down manner, which is significantly faster and yields a compact and easy-to-interpret sequence of modeling instructions.
Our model is also more effective as a shape detector compared to existing state-of-the-art detection techniques.
We finally demonstrate that our network can be trained on novel datasets without
ground-truth program annotations through policy gradient techniques.

\end{abstract}

\section{Introduction}
\vspace{-1mm}
In recent years, there has been a growing interest 
in generative models of 2D or 3D shapes, especially through the use of deep neural networks as image or shape priors~\cite{kulkarni2015deep,Choy,Dosovitskiy2017,FanSG17}. 
However, current methods are limited to the generation of low-level shape representations consisting of pixels, voxels, or points. 
Human designers, on the other hand, rarely model shapes as a collection of these individual elements. 
For example, in vector graphics modeling packages (Inkscape, Illustrator, and so on), shapes are often created through higher-level primitives, such as parametric curves (e.g., Bezier curves) or basic shapes (e.g., circles, polygons), as well as operations acting on these primitives, such as boolean operations, deformations, extrusions, and so on. The reason for choosing higher-level primitives is not incidental. 
Describing shapes with as few as possible primitives and operations is highly desirable for designers since it is compact, makes subsequent editing easier, and is perhaps better at capturing aspects of human shape perception such as view invariance, compositionality, and symmetry~\cite{biederman1987recognition}.


\begin{figure}[!htbp]
\centering
\includegraphics[width=0.9\linewidth]{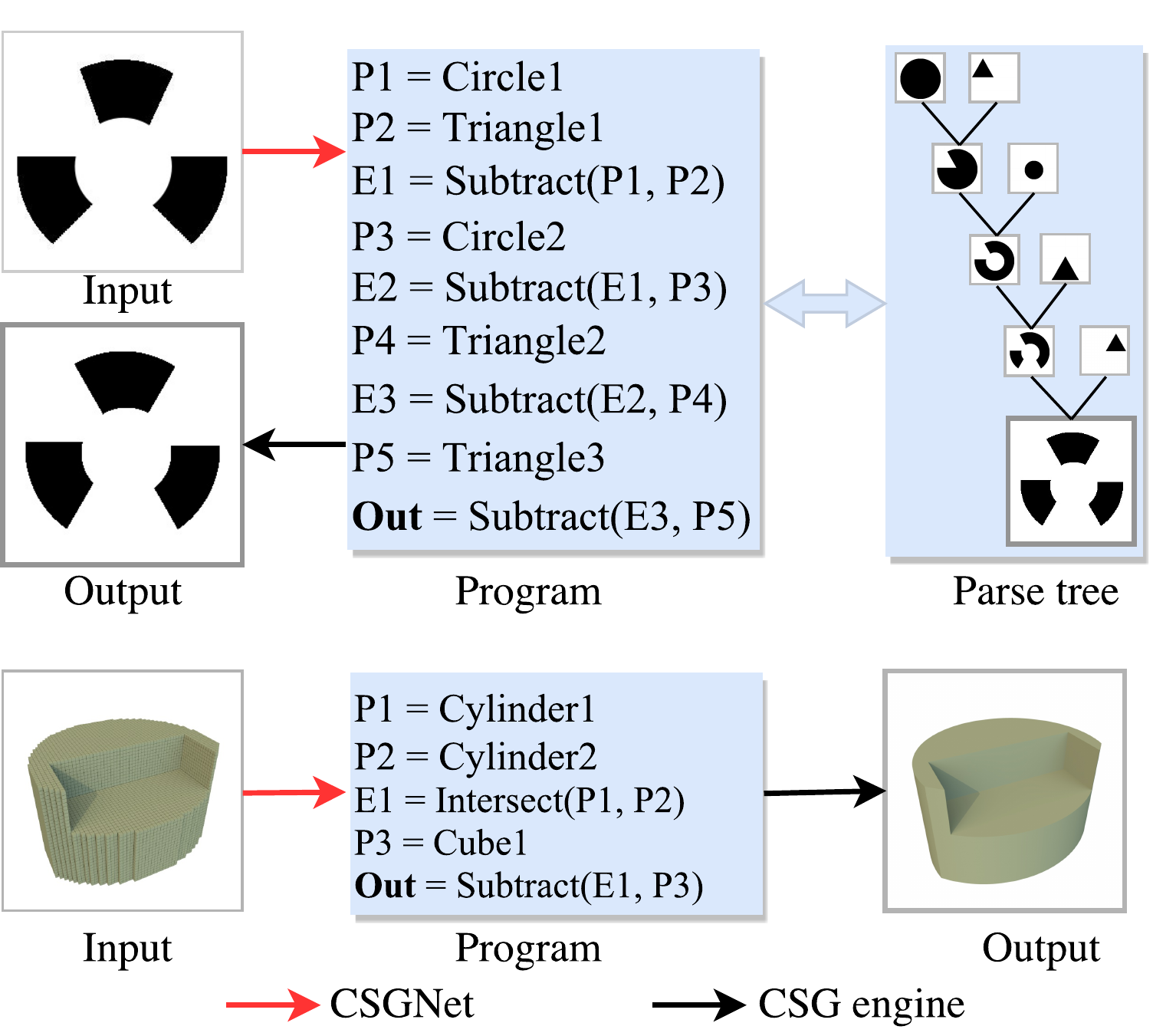}
\caption{\label{fig:teaser}\textbf{Our shape parser produces a compact program that generates an input 2D\ or 3D\ shape}. On top is an input image of 2D\ shape, its program and the underlying parse tree where primitives are combined with boolean operations. On the bottom is an input voxelized 3D\ shape, the induced program, and the resulting shape from its execution.}
\vskip -5mm      
\end{figure}

The goal of our work is to develop an algorithm that parses shapes into their constituent modeling primitives and operations within the framework of Constructive Solid Geometry (CSG) modeling~\cite{Laidlaw1986CSG} as seen in Figure~\ref{fig:teaser}.
This poses a number of challenges.
First, the number of primitives and operations is not the same for all shapes i.e., our output does not have constant dimensionality, as in the case of pixel arrays, voxel grids, or fixed point sets. 
Second, the order of these operations matters. 
Figure \ref{fig:teaser} demonstrates an example where a complex object is created through boolean operations that combine simpler objects. 
If one performs a small change e.g., swap two operations, the resulting object becomes entirely different. 
From this aspect, the shape modeling process could be thought of as a \emph{visual program} i.e., an ordered set of modeling instructions.
Finally, a challenge is that we would like to learn an \emph{efficient} parser that generates a compact program (e.g., with the fewest instructions) without relying on a vast number of shapes annotated with their programs for a target domain.



To tackle these challenges we designed a memory-enabled network architecture, that given a target 2D image of a shape, or a target 3D\ shape, generates a CSG program to generate it. 
To train our network we created a large synthetic dataset of automatically generated 2D and 3D programs.
Networks trained on this dataset however lead to poor generalization when applied to new domains. 
To adapt models to new domains without program annotations we employ policy gradient techniques from the reinforcement learning literature~\cite{Williams92simplestatistical}.
Combining our parser with a CSG rendering engine allows the parser to receive feedback based on the visual difference between the target shape and generated shape.
Thus the parser network can be trained to minimize this difference. 



Our contributions are as follows. First we show that the proposed architecture is efficient and effective at inferring CSG programs for 2D and 3D shapes across a number of domains. Second we show that the parser can be learned using reinforcement learning techniques on novel datasets without program annotations. Third, we show that the parser is a better and faster shape detector than state-of-the art detection approaches that only rely on bottom-up cues. We conjecture that this is because the parser jointly reasons about presence and ordering during parsing unlike the detector. 
\vspace{-1mm}
\section{Related Work}
\begin{figure*}[!htbp]
\centering
\includegraphics[width=\textwidth]{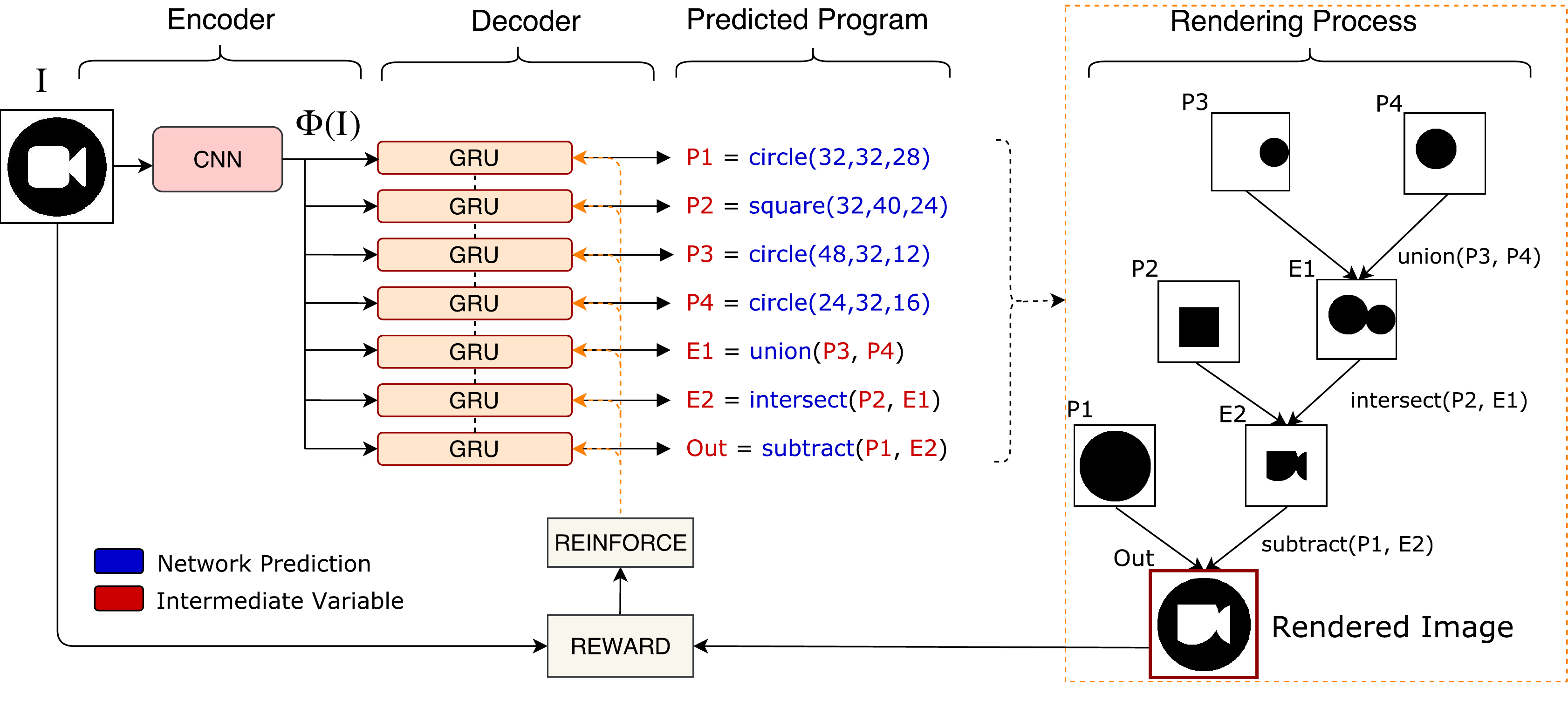}
\vskip -2mm    
\caption{\label{fig:arch}\textbf{Architecture of our neural shape parser (CSGNet)}. CSGNet consists of three parts,
first an encoder takes a shape (2D or 3D) as input and outputs a feature vector through a CNN. 
Second, a decoder maps these features to a sequence of modeling instructions yielding a visual program. 
Third, the rendering engine processes the program and outputs the final shape. The primitives annotated as $P1,P2,P3,P4$  are predicted by the network, while $E1,E2$ are the outputs of boolean modeling operations acting on intermediate shapes.}
\vskip -3mm
\end{figure*}

\vspace{-2mm}
Our work is primarily related to neural program induction methods. Secondly, it
is also related to ``vision-as-inverse-graphics'' approaches, as well as neural network-based methods that predict shape primitives or parameters of procedural graphics models. Below, we briefly overview these prior methods, and explain differences from our work.  \paragraph{Neural program induction.} 
Our method is inspired by recent progress in neural network-based methods that
infer programs expressed in some high-level language to solve a task. These
methods often employ variants of recurrent neural networks whose parameters are
trained to predict desired program outputs given exemplar inputs, such as
answers to questions involving complex arithmetic, logical, or semantic parsing
operations
\cite{Neelakantan,Reed2015,Denil,Balog,Joulin2015a,Zaremba2014,Zarembaa,Kaiser2015,Liang2016}.

In the context of visual reasoning, several authors~\cite{Johnson,hu2017learning} proposed architectures that produce programs composed of functions that perform compositional reasoning on the input image. 
They also incorporate an execution engine that produces the result of the program through a neural module network~\cite{Andreas2016NeuralMN}. 
In contrast our method aims to produce a generative program consisting of shape modeling functions that match a target image. 

\paragraph{Vision-as-inverse-graphics.} A well-known approach to  visual analysis is to generate and fit hypotheses of scenes or objects to input  image data i.e., perform analysis-by-syntesis \cite{yuillek06}.
Kulkani \etal \cite{Kulkarni2015inverse} proposed sampling-based probabilistic inference to estimate parameters of stochastic graphics models (e.g., human body parameters, or parameters of rotationally symmetric objects)  representing the space of hypothesized scenes  given an input image.
Shape grammars (or so-called inverse procedural modeling techniques) have alternatively been used in analysis-by-synthesis image parsing frameworks \cite{TeboulKSKP11,Martinovic:2013:BGL}, yet they have the disadvantage of not modeling long-range dependencies in the parsing task, and are often specific to a particular shape class (e.g., buildings).
More recent approaches employ Convolutional Neural Network (CNN) to infer parameters of objects \cite{kulkarni2015deep} or whole scenes \cite{Romaszko2017inverse}. A similar
trend is observed in graphics applications where CNNs are used to map  input images or partial shapes
to procedural model parameters \cite{huang2017shape,RitchieTHG16,Nishida:2016:ISU}. Wu \etal \cite{Wu} detect objects in scenes by employing a network for producing object proposals  and a network that predicts whether there is an object in a  proposed segment, along with various object attributes.
Eslami \etal \cite{Eslami16} uses a
recurrent neural network to
attend
to one object at a time in a scene,
and learn to use an
appropriate number
of inference steps to recover object counts,
identities and  poses. 

In contrast, we do not aim at parsing images or scenes into a collection of objects and their parameters. We 
 instead parse input images or 3D\ shapes into a sequence of  modeling operations on primitives (i.e, a visual program) to match a target
image. In our setting, the space of outputs is much larger and the order of
operations in our visual programs matter. To deal with this 
complexity, we use a combination of supervised pretraining, reinforcement learning, reward design, and post-optimization of
modeling parameters, described in the next Section.

\paragraph{Neural primitive fitting.} Tulsiani \etal
\cite{abstractionTulsiani17} proposed a volumetric convolutional network
architecture that predicts a fixed number of cuboidal primitives to describe an
input 3D shape. To better handle a variable number of primitives, Zou \etal
\cite{Zou} instead proposed an LSTM-based architecture that predicts boxes given
input depth images. We also aim at deriving geometrically interpretable
explanations of shapes in terms of primitives. However, our network is not limited to predicting
a single type of primitives (e.g., cubes), but also outputs modeling operations acting on
them, or in other words supports a significantly richer  modeling
paradigm. The program can be used not only to geometrically describe the input
shape but can also be directly edited to manipulate it if desired. Finally, Ellis \etal
\cite{Ellis} proposed a neural network architecture to extract various
hand-drawn primitives (lines, circles, rectangles) in images, which are then
grouped into Latex programs. Their program synthesis is posed as a constraint
satisfaction problem which is computationally\ expensive and can take hours to
solve. Instead, our program is created by a neural network that takes a fraction of a second to evaluate at test time.

\paragraph{Bottom-up parsing.} Our work is related to approaches for shape parsing using grammars~\cite{fischler1973representation,felzenszwalb2005pictorial,yang2011articulated,bourdev2010detecting,Bokeloh:2010:CPS,TeboulKSKP11,Martinovic:2013:BGL,Talton:2012:LDP,ritchie2015controlling}. 
These have been applied to objects  that can be represented using tree-structured grammars (e.g., human bodies, buildings). 
However such approaches often use shallow grammars or accurate bottom-up proposals (e.g., face and limb detection) to guide parsing. 
In the context of CSG, primitive detection is challenging as shapes  change significantly when boolean operations are applied to them. Parse trees for CSG\ also tend to be deeper. 
As a result, bottom-up parsing becomes computationally expensive since the complexity scales exponentially with the program length.

\vspace{-1mm}
\section{Designing a Neural Shape Parser} \label{shape-parsing}
\vspace{-2mm}
In this section, we first present our neural shape parser that can induce programs for 2D/3D shapes. The goal of the \textbf{parser} $\pi$ is to produce a sequence of instructions given an input shape.
The parser can be implemented as an encoder-decoder using  neural network modules as shown in Figure~\ref{fig:arch}.
The \textbf{encoder} takes as input an image $I$ and produces an encoding $\Phi(I)$ using a CNN. The \textbf{decoder} $\Theta$ takes as input $\Phi(I)$ and produces a probability distribution over programs
$P$ represented as a sequence of instructions. 
Decoders can be implemented using Recurrent Neural Networks (RNNs).
We employ Gated Recurrent Units (GRUs)~\cite{chung2014empirical} that have been widely used for sequential prediction tasks such as generating natural language and speech. The overall network can be written as $\pi(I) = \Theta \circ \Phi (I)$. 
The space of programs can be efficiently described according to a context-free grammar~\cite{hopcroft}. 
For example, in constructive solid geometry the instructions consist of drawing primitives (\eg,
spheres, cubes, cylinders, \etc) and performing boolean operations described as a grammar with the following production rules:
\vspace{-1mm}
\begin{align*}
&S \rightarrow E \\
&E \rightarrow E~E~T ~|~ P \\
&T \rightarrow \texttt{OP}_1 |\texttt{OP}_2 | \ldots |  \texttt{OP}_m  \\
&P \rightarrow \texttt{SHAPE}_1|\texttt{SHAPE}_2| \ldots | \texttt{SHAPE}_n
\end{align*}
Each rule indicates possible derivations of a non-terminal symbol separated by the $|$ symbol.
Here $S$ is the start symbol, $\texttt{OP}_i$ is chosen from a set of defined  modeling operations and the $\texttt{SHAPE}_i$ is a primitive chosen from a set of basic shapes at different positions, scales, orientations, etc.
Instructions can be written in a standard post-fix notation, \eg $\texttt{SHAPE}_1\texttt{SHAPE}_2\texttt{OP}_1\texttt{SHAPE}_3\texttt{OP}_2$. Figure~\ref{fig:arch} also gives an example of a program predicted by the network, that follows the grammar described above.

\subsection{Learning}\label{learning}
\vspace{-2mm}
Given an input $I$ the parser network $\pi$ generates a program that minimizes a reconstruction error between the shape produced by executing the program and a
target shape. Note that not all programs are valid hence the network must also
learn to generate grammatical programs.

\paragraph{Supervised learning:} \label{supervised-learning}
When target programs are available the architecture can be trained with standard
supervised learning techniques. 
Training data in this case consists of shape and
program pairs $(I^i, P^i), i=1, \ldots, N$. 
In our implementation, the RNN
produces a categorical distribution $\pi_{\theta}$ over instructions $a \in A$
at every time step. 
Similarly the ground-truth program $P^i$ can be written as
sequence of instructions $g^i_1$, $g^i_2$ .. $g^i_{T_i}$, where $T_i$ is the
length of the program $P^i$. The parameters $\theta$ can be learned to maximize the
log-likelihood of the ground truth instructions:
  \begin{equation}
    {\cal L}(\theta) = \sum_{i=1}^{N} \sum_{t=1}^{T_i} \log \pi_\theta(g^i_t|g_{1:t-1},I^i).
  \end{equation}


\paragraph{Learning with policy gradients.} \label{policy-gradient} Without
  target programs one can minimize a reconstruction error between the shape
  obtained by executing the program and the target. 
  However, directly
  minimizing this error using gradient-based techniques is not possible since the output space is discrete and
  execution engines are typically not differentiable. 
  Policy gradient techniques~\cite{Williams92simplestatistical} from the
  reinforcement learning (RL) literature can instead be used in this case.
  

  Concretely, the parser $\pi_{\theta}$, that represents a policy network, can
  be used to sample a program $y$ = ($a_1$,$a_2$ .. $a_T$) conditioned on the
  input shape $I$. Then a reward $R$  can be estimated by measuring
  the similarity between the generated image $\hat{I}$ obtained by executing the program and the target shape $I$. With this setup, we want to learn the network parameters $\theta$ that maximize the expected
  rewards over programs sampled under the predicted distribution $\pi_{\theta}(I)$ across images $I$ sampled from a distribution ${\cal D}$:
\begin{equation*} \label{expected-return}
  \mathbb{E}_{I \sim {\cal D}} \left[ J_\theta(I) \right] = \mathbb{E}_{I \sim {\cal D}} \mathbb{E}_{y\sim \pi_\theta(I))}\left[ R\right].
\end{equation*}
The outer expectation can be replaced by a sample estimate on the training data.
The gradient of the inner expectation can be obtained by rearranging the
equation as:

\begin{equation*} \label{expected-grad}
  \nabla_{\theta} J_\theta(I)= \nabla_{\theta} \sum_y \pi_\theta(y) R  = \sum_y \nabla_{\theta} \log \pi_\theta(y) \left[\pi_\theta(y) R \right].
\end{equation*}

It is often intractable to compute the expectation $J_\theta(I)$ since the
space of programs is very large. Hence the expectation must be approximated.
The popular REINFORCE~\cite{Williams92simplestatistical} algorithm computes a
Monte-Carlo estimate as:
\begin{equation*}
  \nabla_{\theta} J_\theta(I) = \frac{1}{S} \sum_{s=1}^{S} \sum_{t=1}^{T} \nabla \log \pi_\theta (\hat{a}_t^s| \hat{a}_{1:t-1}^s, I)R^s, 
\end{equation*}
by sampling $S$ programs from the policy $\pi_{\theta}$. Each program $y^s$ is
obtained by sampling instructions $\hat{a}^s_{t=1:T}$ from the distribution
$\hat{a}_t^s\sim\pi_{\theta}(a_t|\hat{a}^s_{1:t-1}; I)$ at every time step $t$,
till the stop symbol (\texttt{EOS}) is sampled. The reward $R^s$ is calculated by executing the
program $y^s$. Sampling-based estimates typically have high variance that can be
reduced by subtracting a baseline without changing the bias as:
\begin{equation}
  \nabla_{\theta} J_\theta(I) = \frac{1}{S} \sum_{s=1}^{S} \sum_{t=1}^{T} \nabla_{\theta} \log \pi_{\theta} (\hat{a}_{t}^s| \hat{a}_{1:t-1}^s, I)(R^s - b).
\end{equation}
A good choice of the baseline is the expected value of returns starting
from $t$ \cite{Sutton:1999:PGM:3009657.3009806,Williams92simplestatistical}. We compute baseline as the running average of past rewards.

\paragraph{Reward.} The rewards should be primarily designed to encourage visual similarity of the generated program with the target. Visual similarity between two shapes is measured using the Chamfer distance (CD) between points on the edges of each shape.
The CD is between two point sets, $\mathbf{x}$ and $\mathbf{y}$, is defined as follows:
\begin{equation}
        Ch(\mathbf{x}, \mathbf{y})=
                 \frac{1}{2|\mathbf{x}|}\sum_{x \in \mathbf{x}} \min_{y \in \mathbf{y}} \norm{x-y}_2 +
                 \frac{1}{2|\mathbf{y}|}\sum_{y \in \mathbf{y}} \min_{x \in \mathbf{x}} \norm{x-y}_2. \nonumber
\end{equation}
The points are scaled by the image diagonal, thus   $Ch(\mathbf{x}, \mathbf{y}) \in [0,1]~\forall \mathbf{x}, \mathbf{y}$. 
The distance can be efficiently computed using distance transforms.
In our implementation, we also set  a maximum length $T$ for the induced programs to avoid having too long or  redundant programs (e.g., repeating the same modeling instructions  over and over again). We then define the reward as:
\begin{equation*}
  R =  \begin{cases}
    f\left(Ch(\texttt{Edge}(I),\texttt{Edge}(Z(y)\right), &y \text{ is valid} \\
    0,          &y \text{ is invalid}. \\
    \end{cases}
  \end{equation*}
where $f$ is a shaping function and $Z$ is the CSG rendering engine.
Since invalid programs get zero reward, the maximum length constraint on the programs
encourages the network to produce shorter programs with high rewards. We use maximum length $T=13$ in all of our RL experiments. The function $f$ shapes the CD as
  $f(x) = (1-x)^\gamma$ with an exponent $\gamma > 0$.
  Higher values of $\gamma$ encourages CD close to zero. 
  We found that $\gamma=20$ provides a good trade-off between program length and visual similarity.
  


\subsection{Inference} \label{inference}
\vspace{-2mm}
\paragraph{Greedy decoding and beam search.} Estimating the most likely program given an input is intractable using RNNs.
Instead one usually employs a greedy decoder that picks the most likely instruction at each time step.
An alternate is to use a beam search procedure that maintains the k-best likely
sequences at each time step. In our experiments we report results with varying beam sizes.

\paragraph{Visually-guided refinement.}
Our parser produces a program with a discrete set of primitives. However, further refinement can be done by directly optimizing the position and size of the primitives to maximize the reward. 
The refinement step keeps the program structure of the program and primitive type fixed but uses a heuristic algorithm~\cite{Powell1964} to optimize the parameters using feedback from the rendering engine.
On our dataset where shapes have up to $7$ primitives, the search space is relatively small and the algorithm converges to a local minima in about $10$ iterations and consistently improves the results.

\vspace{-1mm}
\section{Experiments} \label{experiment}
\vspace{-2mm}
We describe our experiments on different datasets exploring the generalization capabilities of our network (CSGNet).
We first describe our datasets: (i)\ an automatically generated dataset of 2D and 3D\ shapes based on synthetic generation of CSG\ programs, (ii) 2D\ CAD\ shapes mined from the web where ground-truth programs are not available, and (iii) logo images mined also from the web where ground-truth programs are also not available. We discuss our qualitative and quantitative results on the above  datasets. 

\vspace{-1mm}
  \subsection{Datasets} \label{grammar}
\vspace{-2mm}  
\begin{figure}[h]
        \centering
        \includegraphics[width=\linewidth]{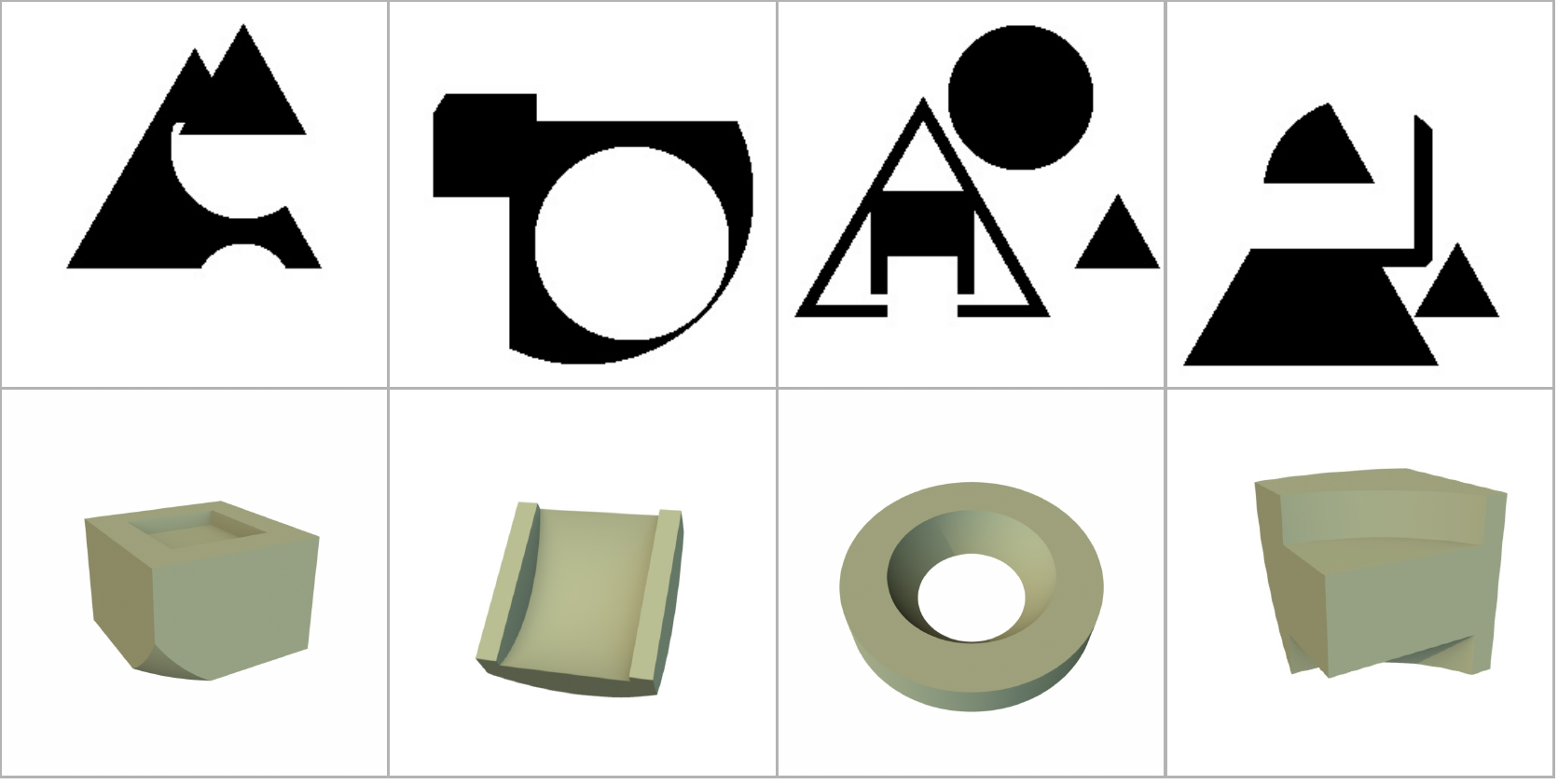}
        \caption{\textbf{Samples created from our synthetically generated programs.} 2D\ samples in top row and 3D\ samples in bottom row.}
        \label{fig:samples}
\end{figure}
To train our network in the supervised learning setting, we automatically created a large set of 2D\ and 3D CSG-based synthetic programs according to the  grammars described below. \paragraph{Synthetic 2D shapes.}
We sampled derivations of the following CSG grammar to create our synthetic dataset in the 2D\ case:
\vspace{-2mm}
\begin{align*}
&S \rightarrow E;\\
&E \rightarrow EET \mid P(L, R);\\
&T \rightarrow intersect \mid union \mid subtract;  \\
&P \rightarrow square \mid circle  \mid triangle;  \\
&L \rightarrow \big[8:8:56\big]^2;~~ R \rightarrow \big[8:4:32\big].
\end{align*}
Primitives are specified by their type: \emph{square}, \emph{circle}, or
\emph{triangle}, locations $L$ and circumscribing circle of radius $R$ on a
canvas of size $64\times64$. There are three boolean operations: $intersect$, $union$, and $subtract$. L is discretized to lie on a square grid with spacing of $8$ units and R is discretized with spacing of $4$ units. The \emph{triangles} are assumed to be upright and equilateral. The synthetic dataset is created by sampling random programs containing different number of primitives from the above grammar, constraining the distribution of various primitive types and operation types to be uniform. We also ensure that no duplicate programs exist in our dataset. The primitives are rendered as binary images and the programs are executed on a canvas of $64\times64$ pixels. Samples from our dataset are shown in Figure \ref{fig:samples}. Table \ref{table:dataset} provides details about the size and splits of our dataset.
\begin{table}
  \centering
  \setlength{\tabcolsep}{3pt}
\begin{tabular}{c|c|c|c|c|c|c}
\multirow{2}{*}{\begin{tabular}[c]{@{}c@{}}Program\\ Length\end{tabular}} & \multicolumn{3}{c|}{2D} & \multicolumn{3}{c}{3D} \\
\cline{2-7}
                              & Train   & Val  & Test  & Train   & Val  & Test  \\
\hline
3                             & 25k     & 5k   & 5k    & 100k    & 10k  & 20k   \\
5                             & 100k    & 10k  & 50k   & 200k    & 20k  & 40k   \\
7                             & 150k    & 20k  & 50k   & 400k    & 40k  & 80k   \\
9                             & 250k    & 20k  & 50k   & -       & -    & -     \\
11                            & 350k    & 20k  & 100k  & -       & -    & -     \\
13                            & 350k    & 20k  & 100k  & -       & -    & -    
\end{tabular}
\vskip 1mm
\caption{\textbf{Statistics of our 2D and 3D synthetic dataset.}}
  \label{table:dataset}
\vskip -2mm    
\end{table}

\paragraph{Synthetic 3D shapes.}
We sampled derivations of the following  grammar  in the case of 3D\ CSG:
\vspace{-2mm}
\begin{align*}
  &S \rightarrow E; ~E \rightarrow EET; \\
  &E \rightarrow sp(L,R) \mid cu(L,R) \mid cy(L,R,H)\\
  &T \rightarrow intersect \mid union \mid subtract;  \\
  &L \rightarrow \big[8:8:56]^3\\
  &R \rightarrow \big[8:4:32];~H \rightarrow \big[8:4:32].
\end{align*}
The same three binary operations are used as in the 2D\ case. Three basic solids are denoted by
`$sp$': Sphere, `$cu$': Cube, `$cy$': Cylinder. $L$ represents the center of
primitive in 3D voxel grid. $R$ specifies radius of sphere and cylinder, and
also specifies size of cube. $H$ is the height of cylinder. The primitives are
rendered as voxel grids and the programs are executed on a 3D volumetric grid of size $64$ $\times$ $64$ $\times$ $64$. We used the same random sampling method as described
for the synthetic 2D dataset, resulting in 3D\ CSG\ programs.  3D\ shape samples from this dataset are shown in Figure
\ref{fig:samples}.

\paragraph{2D CAD shapes.}
We collected $8K$ CAD shapes from the Trimble 3DWarehouse dataset~\cite{TrimbleWarehouse}
in three categories: chair, desk and lamps. We rendered the
CAD\ shapes into $64$ $\times$ $64$ binary masks from their front and side views. In
Section \ref{experiment}, we show that the rendered shapes can be parsed
effectively through our visual program induction method. We split this dataset into $5K$ shapes for  training, $1.5K$ validation and $1.5K$ for testing.
\paragraph{Web logos.}
We mined  a  collection of binary logos  from the web that can be modeled using the primitives in our output shapes. We test our approach on these logos without further training or fine-tuning  our net on this data.

\subsection{Implementation details}
\vspace{-1mm}
The input 2D or 3D shape $I$ is represented as pixel and voxel occupancy grid
respectively. Our encoder is based on an image-based convnet in the case of 2D\
inputs, and a volumetric convnet in the case of 3D\ inputs. The output of the
encoder $\Phi(I)$ is passed as input to our GRU-based decoder at every program
step. The hidden state of our GRU units is passed through two fully-connected
layers, which are then converted into a probability distribution over program
instructions through a classification layer. For the 2D CSG there are $400$
unique instructions corresponding to $396$ different primitive types, discrete
locations and sizes, the $3$ boolean operations and the stop symbol. For the 3D
CSG there are $6635$ unique instructions with $6631$ different types of
primitives with different sizes and locations, plus $3$ boolean modeling operations and a stop symbol. During training, on synthetic dataset, we sample images rendered from programs of variable length (up to $13$ for 2D and up to $7$ for 3D dataset) from training dataset. More details about the architecture
of our encoder and decoder (number and type of layers) are provided in the
supplementary material.


For supervised learning, we use the Adam optimizer~\cite{KingmaB14}
with learning rate $0.001$ and dropout of $0.2$ in non-recurrent network
connections. For reinforcement learning, we use stochastic gradient descent
with $0.9$ momentum, $0.01$ learning rate, and with the same dropout as above.
Our implementation is based on PyTorch \cite{pytorch}. Our source code and datasets are available on our project page: \url{https://hippogriff.github.io/CSGNet}.

\subsection{Results}
\vspace{-1mm}
We evaluate our network, called CSGNet, in two different ways: (i) as a model for inferring the entire program, and (ii) as model for inferring  primitives, i.e., as an object detector.

\vspace{-4mm}
\subsubsection{Inferring programs}
\begin{figure}[!htbp]
\centering
\includegraphics[width=\linewidth]{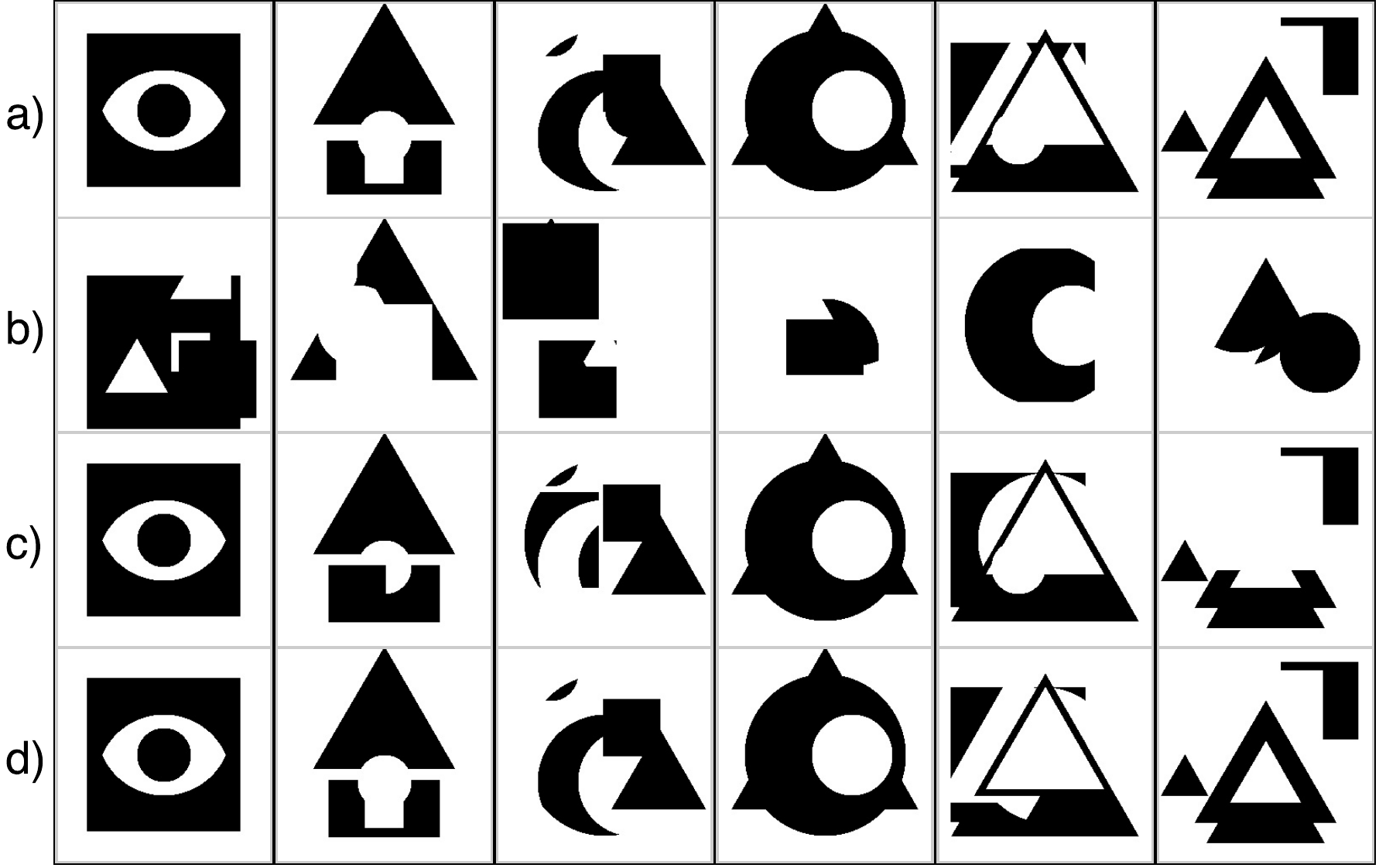}
\caption{\textbf{Comparison of performance on synthetic 2D dataset.} a) Input image, b) NN-retrieved image, c) top-$1$ prediction, and d) best result from top-$10$ beam search predictions of CSGNet.}
  \vskip -3mm
\label{fig:synth2d-comparison}
\end{figure}

\vspace{-2mm}
\paragraph{Evaluation on the synthetic 2D shapes.} \label{synthetic-experiment} 
We perform supervised learning to train CSGNet on the training split of this synthetic dataset, and evaluate performance on its test split under different beam sizes. We compare with a baseline that retrieves a program in the training split using a Nearest Neighbor (NN) approach. In NN setting, the program for a test image is retrieved by taking the program of the train image that is most similar to the test image.
Table~\ref{2D-dataset-table} compares CSGNet to this NN\ baseline using the Chamfer distance between the test target and predicted shapes. 
Our parser is able to outperform the NN method. One would expect that NN\ would perform well here because the size
of the training set is large. However, our results indicate that our compositional parser
is  better at capturing  shape variability, which is still significant in this dataset. Results are
also shown with increasing beam sizes (k) during decoding, which consistently
improves performance. Figure~\ref{fig:synth2d-comparison} also shows the programs retrieved through NN
and our generated program for a number of characteristic examples in our test  split of our synthetic dataset.

\begin{table}
\centering
\begin{tabular}{l|c|c|c|c}
        \multirow{2}{*}{Method} & \multirow{2}{*}{NN} & \multicolumn{3}{c}{CSGNet} \\
        \cline{3-5}
        &                     & k=$1$        & k=$5$ & k=$10$      \\
        \hline
        CD                     & $1.94$              & $1.69$       & $1.46$& $\textbf{1.39}$     
\end{tabular}

\vskip 1mm
\caption{\textbf{Comparison of the supervised network (CSGNet) with the NN baseline on the synthetic 2D dataset.} Results are shown using Chamfer Distance (CD) metric by varying beam sizes ($k$) during decoding. CD is in number of pixels.} 
\vskip -3mm
\label{2D-dataset-table}
\end{table}
\paragraph{Evaluation on 2D\ CAD shapes.} \label{cad-experiment}
For this dataset, we report results on its test split under two conditions: (i) when training our network only on  synthetic data, and (ii) when  training our network  on  synthetic data and also fine-tuning it on the training split of 2D\ CAD dataset using policy gradients.

\begin{figure}
\centering
\includegraphics[width=1.0\linewidth]{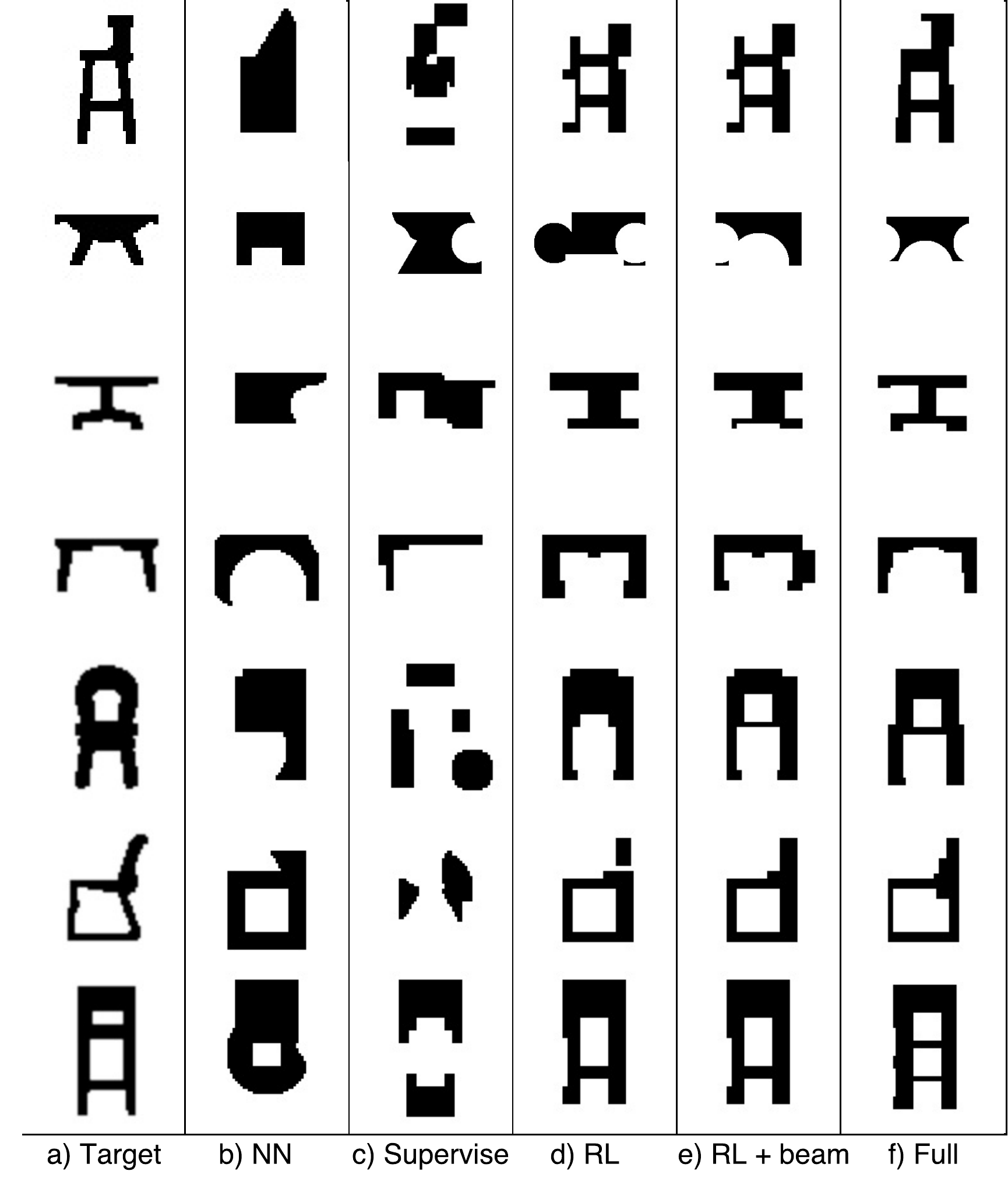}
\caption{\label{fig:cad-results}\textbf{Comparison of performance on the 2D\ CAD dataset}. From left column to right
  column: a) Input image, b) NN retrieved image, c) top-1 prediction from CSGNet in the supervised learning mode, d) top-1 prediction from CSGNet fine-tuned with RL (policy gradient), e) best result from beam search from CSGNet fine-tuned with RL, f) refining our results using the visually guided search on the best beam result (``full'' version of our method).}
\vskip -3mm
\end{figure}

Table~\ref{table:CAD} shows quantitative results on this dataset. We first compare with the NN baseline.
For any shape in this dataset, where ground truth program is not available, NN retrieves a shape from synthetic dataset and we use the ground truth program of the retrieved synthetic shape for comparison. We then list the performance of CSGNet trained in supervised manner only on our synthetic dataset. With beam search, the performance of this variant improves compared to NN. Most importantly, further training with Reinforcement Learning (RL) on the training split of the 2D\ CAD\ dataset improves the results significantly and outperforms the NN approach by a considerable margin. 
This also shows the advantage of using RL, which trains the shape parser without ground-truth programs. We  note that directly training the network using RL alone does not yield good results  which suggests that the two-stage learning (supervised learning and RL) is important. Finally,  optimizing the best beam search  program with visually guided refinement yielded results with the smallest Chamfer Distance.
Figure~\ref{fig:cad-results} shows a comparison of the rendered programs for various examples in the test split of the 2D\ CAD\ dataset for variants of our network. Visually guided refinement on top of beam search of our two stage-learned network qualitatively produces results that best match the input image.
\begin{table}[]
        \centering
        \setlength{\tabcolsep}{3pt}
        \begin{tabular}{l|c|c|c|c|c|c|c|c}
                \multirow{2}{*}{Method} & \multirow{2}{*}{Train} & \multirow{2}{*}{Test} & \multicolumn{6}{c}{CD (@refinement iterations)} \\
                \cline{4-9}
                &                        &                       & $i$=$0$      & $i$=$1$       & $i$=$2$       & $i$=$4$       & $i$=$10$      & $i$=$\infty$\\
                \hline
                NN            & -                      & -                     & 1.92 & 1.22 & 1.13 & 1.08  & 1.07  & 1.07  \\
                \hline
                CSGNet        & Super                  &  k=1                     & 2.30 & 1.13 & 0.97   & 0.91 & 0.90  & 0.90   \\
                CSGNet        & Super                  &  k=10                     & 1.60 & 0.71 & 0.60 & 0.56 & 0.55 & 0.55 \\
                \hline
                CSGNet        & RL                     &  k=1                     & 1.26  & 0.61 & 0.54 & 0.52 & 0.51 & $0.51$ \\
                CSGNet        & RL                     &  k=10                     & 1.14 & 0.50 & 0.44 & 0.42   & 0.42  & $\textbf{0.41}$
        \end{tabular}
        \vskip 1mm
        \caption{\textbf{Comparison of various approaches on the CAD shape dataset}. CSGNet trained with supervision (Super) is comparable to the NN approach but reinforcement learning (RL) on the CAD dataset significantly improves the results. Results are shown with different beam sizes ($k$) during decoding. Increasing the number of iterations ($i$) of visually guided refinement during testing improves results significantly. CD metric is in number of pixels.}
        \label{table:CAD}
        \vskip -2mm
\end{table}
\vspace{-1mm}

\begin{figure}[!htbp]
\centering
\includegraphics[width=1.0\linewidth]{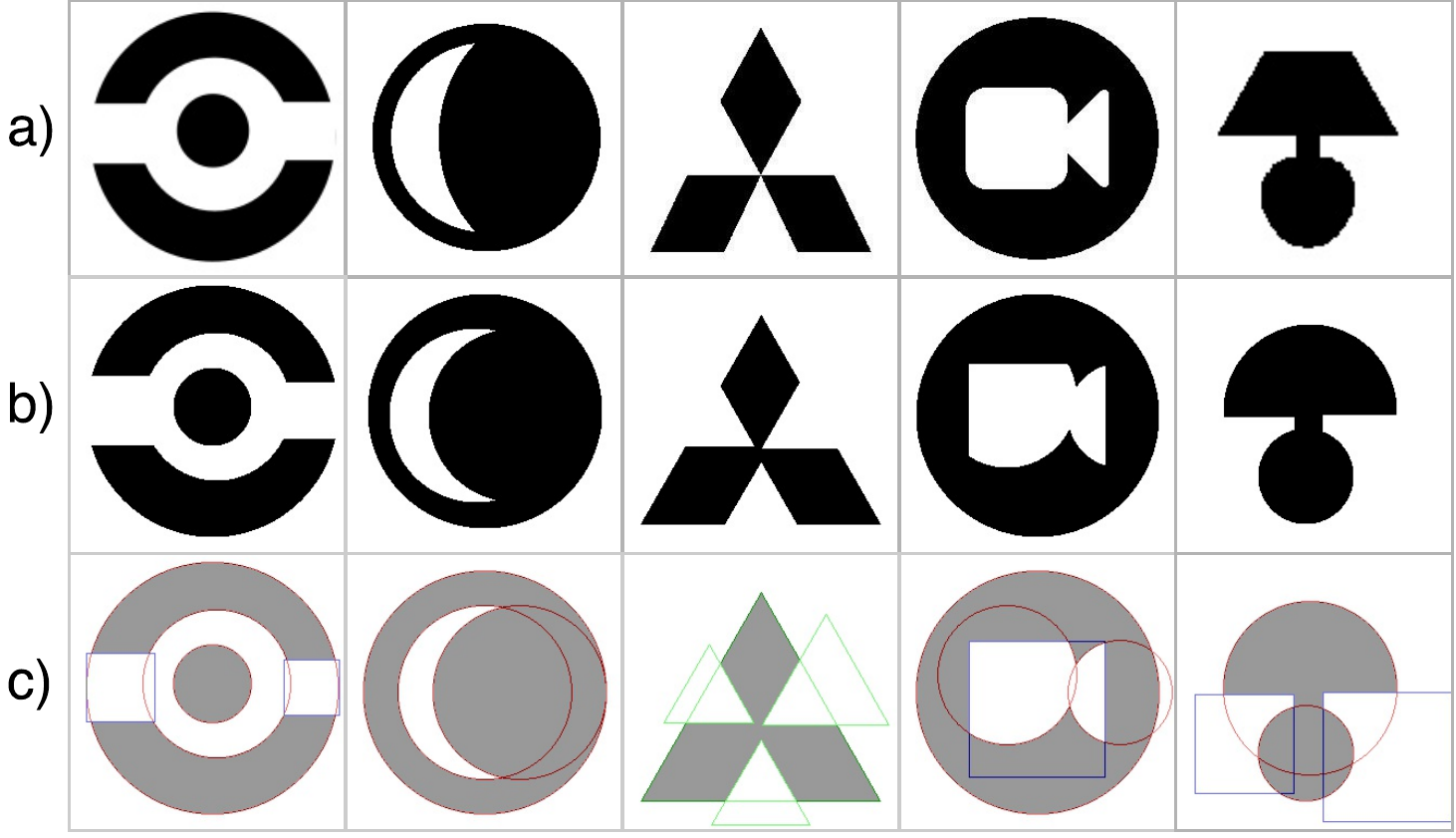}
\vskip 1mm
\caption{\textbf{Results for our logo dataset}. a) Target logos, b) output shapes from CSGNet  and c) inferred primitives from output program. Circle primitives are shown with red outlines, triangles with green  and squares
  with blue.}
\vskip -4mm
\label{fig:logos}
\end{figure}

\paragraph{Logos.}
Here, we experiment with the logo dataset described in  Section
\ref{grammar} (none of these logos participate in training). 
Outputs of the induced programs parsing the input logos are shown in Figure \ref{fig:logos}. In general,
our method is able to parse logos into primitives well, yet performance can degrade when long  programs are required to
generate them, or when they contain shapes that are very different from our used primitives. 

\begin{figure}[!htbp]
\centering
\includegraphics[width=1.03\linewidth]{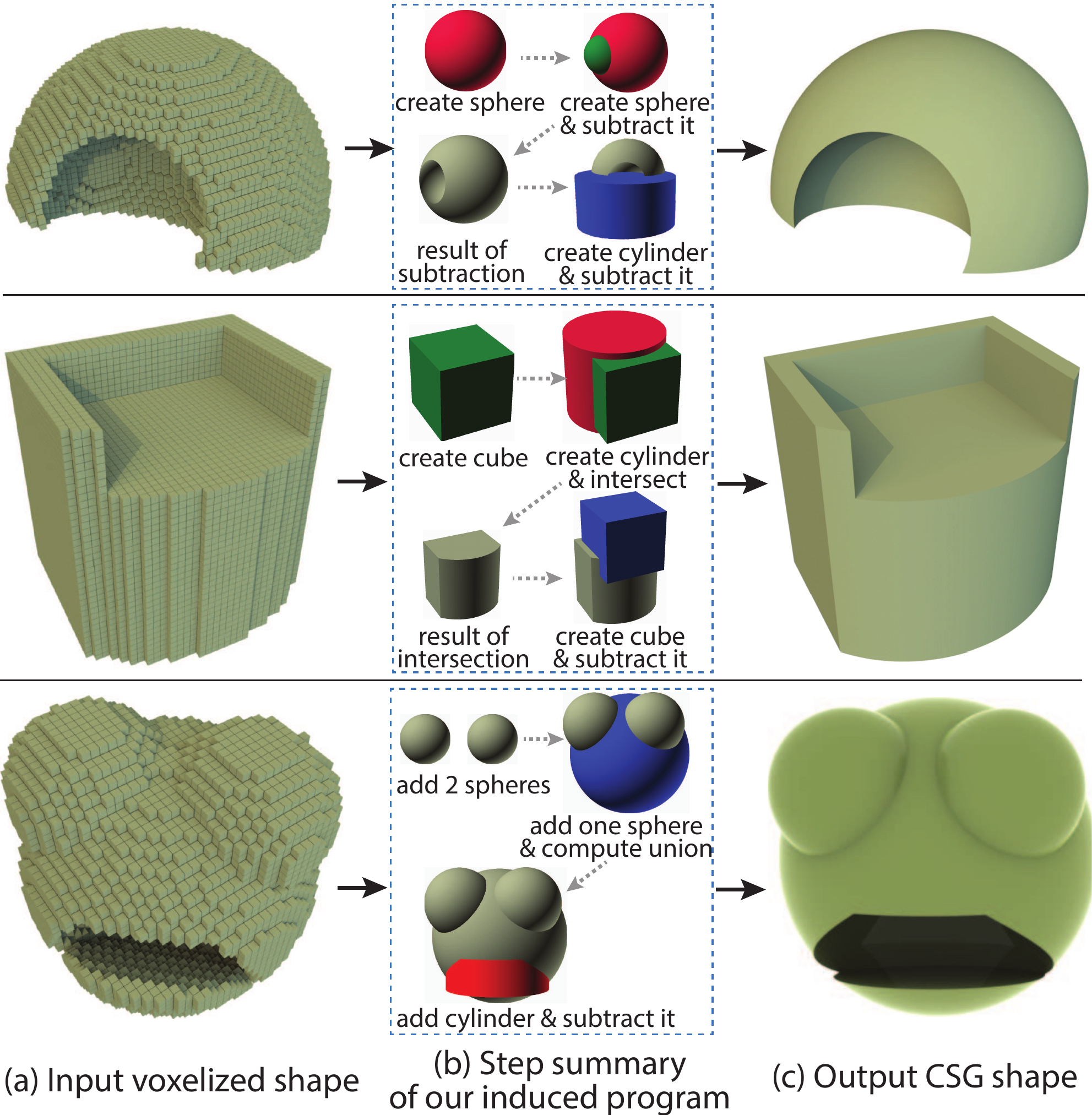}
\vskip 2mm
\caption{\textbf{Qualitative performance of 3D-CSGNet.} a) Input voxelized shape, b) Summarization of the steps of the program induced by CSGNet in the form of intermediate shapes, c) Final output created by executing induced program.}
\vskip -1mm
\label{fig:3D-results}
\end{figure}

\paragraph{Evaluation on Synthetic 3D CSG.} Finally, we show that our approach can  be extended to 3D
shapes. In the 3D\ CSG\ setting, we train a 3D-CNN + GRU (3D-CSGNet) network on  the 3D\ CSG\ synthetic
dataset explained in  Section \ref{grammar}. The input to our 3D-CSGNet  are voxelized shapes in a  $64$$\times$$64$$\times$$64$ grid. Our output is a 3D\ CSG\ program, which can be rendered as a high-resolution polygon mesh (we emphasize that our output is not voxels, but\ CSG\ primitives and operations that can be computed and rendered accurately).  Figure \ref{fig:3D-results}
show  pairs of input voxel grids and our output shapes
from the test split of the 3D\ dataset. The qualitative results are shown in the Table \ref{3D-comparison}, where we compare our 3D-CSGNet at different beam search decodings with NN method. The results indicate that our method is promising in inducing correct programs, which  also have the advantage of accurately reconstructing  the voxelized surfaces into high-resolution surfaces.\  
\begin{table}[]
\centering
\label{3D-comparison}
\setlength{\tabcolsep}{3pt}
\begin{tabular}{l|c|c|c|c}
\multirow{2}{*}{Method} & \multirow{2}{*}{NN} & \multicolumn{3}{c}{3D-CSGNet} \\
\cline{3-5}
&                     & k=$1$        & k=$5$ & k=$10$      \\
\hline
IOU (\%)                     & $73.2$              & $80.1$       & $85.3$& $\textbf{89.2}$     
\end{tabular}
\caption{\textbf{Comparison of the supervised network (3D-CSGNet) with NN baseline on 3D dataset}. Results are shown using IOU(\%) metric by varying beam sizes ($k$) during decoding.}
\label{3D-comparison}
\end{table}

\subsubsection{Primitive detection}
\vspace{-2mm}
Successful program induction for a shape requires not only predicting correct
primitives but also correct sequences of operations to combine these primitives.
Here we evaluate the shape parser as a primitive detector (i.e., we evaluate the output primitives of our program, not the operations themselves).
This allows us to directly compare our approach with bottom-up object
detection techniques.

In particular we compare against a state-of-the-art object detector  (Faster R-CNNs~\cite{NIPS2015-5638}).
The Faster R-CNN is based on the VGG-M network~\cite{Chatfield14} and is trained using bounding-box and primitive annotations based on our 2D\ synthetic training dataset. At test time the detector produces a set of bounding boxes with associated class scores. The models are trained and evaluated on 640$\times$640 pixel images. We also experimented with bottom-up approaches for primitive detection based on Hough transform~\cite{Duda} and other rule-based approaches. However, our  experiments indicated that the Faster R-CNN was considerably better.

For a fair comparison, we obtain primitive detections from CSGNet trained on the 2D\ synthetic dataset only (same as the Faster R-CNN). To obtain detection scores, we sample $k$ programs with beam-search decoding. The primitive score is the fraction of times it appears across all beam programs. This is a Monte Carlo estimate of our detection score.

The accuracy can be measured through standard evaluation protocols for object detection (similar to those in the PASCAL VOC benchmark). We report the Mean Average Precision (MAP) for each primitive type using an overlap threshold between the predicted and the true bounding box of $0.5$ intersection-over-union. Table~\ref{tab:primitive-detection} compares the parser network to the Faster R-CNN approach. 

Our parser clearly outperforms the Faster R-CNN detector on the squares and triangles category. With larger beam search,
we also produce slighly better results for circle detection.
Interestingly, our parser is considerably faster than Faster R-CNN tested on the same GPU.

\begin{table}[!h]
\centering
\setlength{\tabcolsep}{1pt}
\begin{tabular}{l|c|c|c|c|c}
Method     & Circle     & Square     & Triangle     & Mean  & Speed (im/s)\\ \hline
Faster R-CNN   & 87.4 & 71.0 & 81.8 & 80.1 & 5\\ 
CSGNet, $k=10$ & 86.7 & 79.3 & 83.1 & 83.0 & 80\\ 
CSGNet, $k=40$ & \textbf{88.1} & \textbf{80.7} & \textbf{84.1} & \textbf{84.3} & 20
\end{tabular}
\vskip 1mm
\caption{\label{tab:primitive-detection} \textbf{MAP of detectors on the synthetic 2D shape dataset.} We also report detection speed  measured as images/second on a NVIDIA 1070 GPU.}
\label{proposal-comparison}
\end{table}

\vspace{-1mm}
\section{Conclusion}
\vspace{-1mm} We believe that our work represents a first step towards the
automatic generation of modeling programs given target visual content, which we
believe is quite ambitious and hard problem. We demonstrated results of
generated programs in various domains, including logos, 2D\ binary shapes, and
3D\ CAD\ shapes, as well as an analysis-by-synthesis application in the context
of 2D\ shape primitive detection.

One might argue that the 2D\ images and 3D\ shapes our method parsed are
relatively simple in structure or geometry. However, we would also like to point
out that even in this ostensibly simple application scenario (i) our method
demonstrates competitive or even better results than state-of-the-art object
detectors, and most importantly (ii) the problem of generating programs was far
from trivial to solve: based on our experiments, a combination of memory-enabled
networks, supervised and RL strategies, along with beam and local exploration of
the state space all seemed necessary to produce good results. As future work, a
challenging research direction would be to generalize our approach to longer
programs with much larger spaces of parameters in the modeling operations and more 
sophisticated reward functions balancing perceptual similarity to the input image 
and program length.  Other promising directions would be to explore how to combine bottom-up
proposals and top-down approaches for parsing shapes, in addition to exploring top-down program generation strategies.

\textbf{Acknowledgments}. We acknowledge support from NSF (CHS-1422441, CHS-1617333, IIS-1617917) and the MassTech Collaborative grant for funding the UMass GPU cluster.

{\footnotesize
\bibliographystyle{ieee}
\bibliography{egbib}
}
\section{Supplementary}
\paragraph{}
In this supplementary material, we include the following topics in more detail:
a) synthetic dataset creation in the 2D and the 3D case, b) neural
network architecture used in our experiments, c) more qualitative results on
our test dataset.
\subsection{Dataset} \label{dataset}
\begin{figure}
\centering
\includegraphics[width=\linewidth]{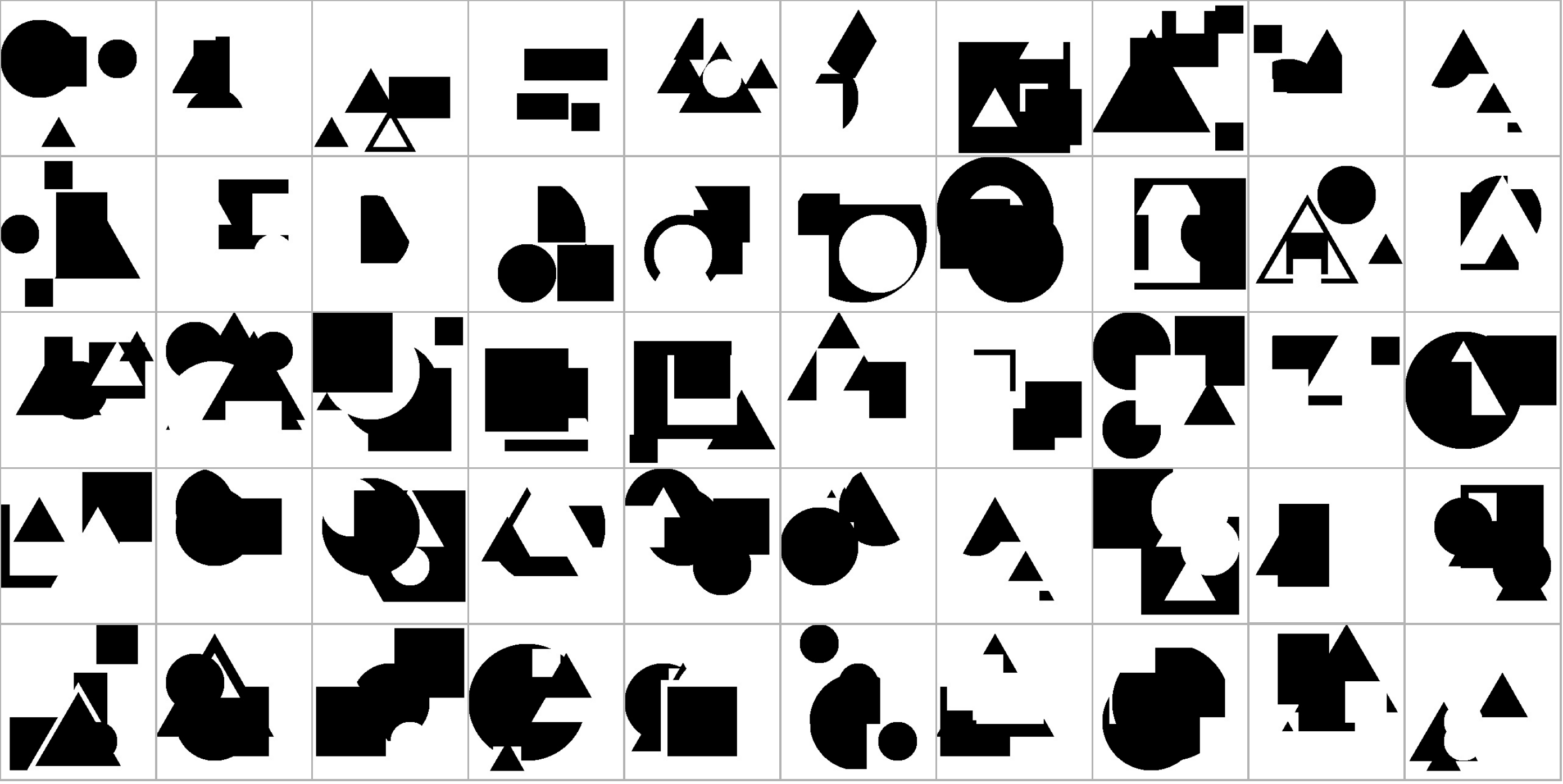}
\caption{\textbf{Random samples from our synthetic 2D dataset.}}
\label{fig:synth2d-examples}
\end{figure}

\paragraph{Synthetic 2D shapes.}
We use the grammar described in the Section $4.1$ to create our 2D dataset. The dataset
is created by randomly generating programs of lengths $3$ to $13$ following the
grammar. While generating these programs we impose additional restrictions as
follows: a) Primitives must lie completely inside the canvas, b) Each operation
changes the number of ON pixels by at least a threshold set to $10\%$ of sum of
pixels in two shapes. This avoids spurious operations such as subtraction
between shapes with little overlap. c) The number of ON pixels in the final
image is above a threshold. d) The previous rules promotes programs with the
$union$ operation. To ensure a balanced dataset we boost the probabilities of
generating programs with $subtract$ and $intersect$ operations. Finally we
remove duplicates. We only use upright, equilateral triangles and
upright squares. Note that locations (L) are discretized to lie on square grid with spacing of $8$ units and size (R) are discretized with spacing of $4$ units. Figure~\ref{fig:synth2d-examples} shows examples from our dataset.

\begin{figure*}[!h]
\centering
\includegraphics[scale=0.35]{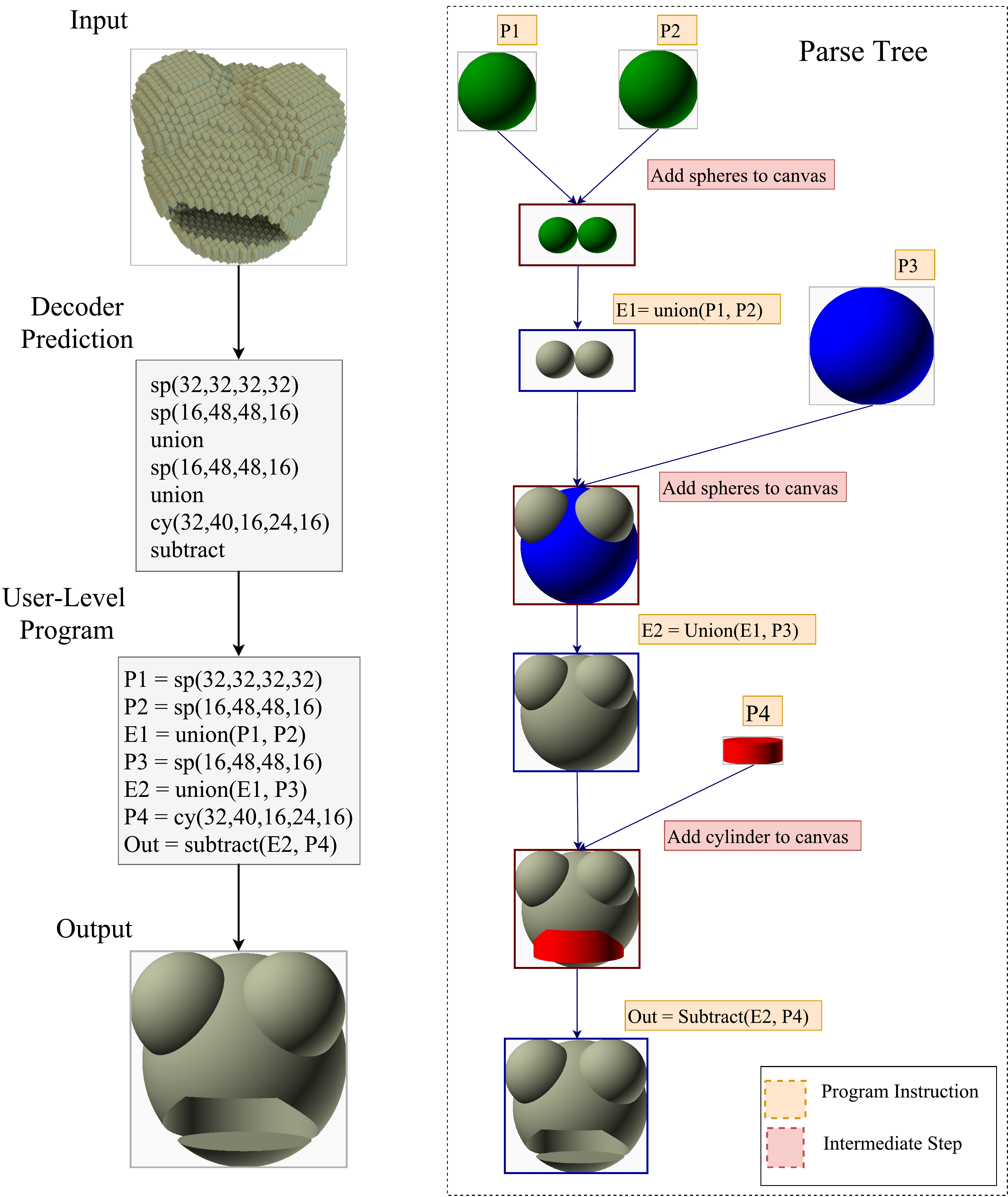}
\caption{\textbf{Detailed execution procedure followed by an induced CSG program in a
  characteristic 3D case.} The input is a voxel based representation of size $64
  \times 64 \times 64$. The RNN decoder produces a program, which can be
  executed following the grammar described in the Section~\ref{dataset}, to give
  the output shown at the bottom. The user-level program is shown for
  illustration. On the right side is shown a parse tree corresponding to the
  execution of the program.}
\label{fig:3d-execution}
\end{figure*}

\paragraph{Synthetic 3D shapes.}
We use the grammar described in the Section $4.1$ to create our 3D dataset. While generating shapes we followed a strategy similar to the 2D case. For 3D case, we
only use programs of up to length $7$ (up to $4$ shape primtives and upto $3$
boolean operations). Note that the cube and cylinder are upright. The dataset
contains $64\times64\times64$ voxel-grid shapes and program pairs. Also note that locations (L) are discretized to lie on cubic grid with spacing of $8$ units, and size (R) and height (H) are discretized with spacing of $4$ units.
\paragraph{CSG execution engine.}
We implemented a CSG engine that reads the instructions one by one. If it
encounters a primitive (e.g. \texttt{c(32, 32, 16)}) it draws it on an empty
canvas and pushes it on to a stack. If it encounters an operation (e.g.
\texttt{union}, \texttt{intersect}, or \texttt{subtract}) it pops the top two
canvases on its stack, applies the operation to them, and pushes the output to
the top of the stack. The execution stops when no instructions remain at which
point the top canvas represents the result. The above can be seen as a set of
shift and reduce operations in a LR-parser \cite{Knuttt}. Figure~\ref{fig:3d-execution} describes execution procedure to induce programs for 3D shapes.

\subsection{Network Architecture}
\begin{table}[t]
	\centering
	\setlength{\tabcolsep}{1pt}
	\begin{tabular}{l|c}
		\textbf{Layers}                                    & \textbf{Output}       \\ \hline
		Input image                                     & $64\times64\times1$  \\ 
		Dropout(Relu(Conv: $3 \times 3$, $1$ $\rightarrow$ $8$))   & $64\times64\times8$  \\ 
		Max-pool($2 \times 2$)                                 & $32\times32\times8$   \\ 
		Dropout(Relu(Conv: $3 \times 3$, $8$ $\rightarrow$ $16$))  & $32\times32\times16$ \\ 
		Max-pool($2 \times 2$)                                 & $16\times16\times16$ \\ 
		Dropout(Relu(Conv: $3 \times 3$, $16$ $\rightarrow$ $32$)) & $16\times16\times32$ \\ 
		Max-pool($2 \times 2$)                               & $8\times8\times32$   \\ 
		Flatten & $2048$
	\end{tabular}
	\vspace{2mm}
	\caption{\textbf{Encoder architecture for 2D shapes experiments.} The same architecture is used in all experiments in the Section $4.3.1$.}
	\label{2dcnnarchitecture}
\end{table}
\begin{table}[]
	\centering
	\begin{tabular}{l|l|c}
		\textbf{Index} & \textbf{Layers}                                    & \textbf{Output}       \\ \hline
		1 & Input shape encoding & $2048$ \\ 
		2 & Input previous instruction & $401$ \\ 
		3 & Relu(FC ($401$ $\rightarrow$ $128$)) & $128$ \\ 
		4 & Concatenate ($1$, $3$) & $2176$ \\ 
		5 & Drop(GRU (hidden dim: $2048$)) & $2048$ \\ 
		6 & Drop(Relu(FC($2048$ $\rightarrow$ $2048$))) & $2048$ \\ 
		7 & Softmax(FC($2048$ $\rightarrow$ $400$)) & $400$
	\end{tabular}
	\vspace{2mm}
	\caption{\textbf{Decoder architecture for 2D shapes experiments.} The same architecture
		is used for all experiments in the Section $4.3.1$. FC: Fully connected dense
		layer, Drop: dropout layer with 0.2 probability. Dropout on GRU are applied on
		outputs but not on recurrent connections.}
	\label{2d-rnnarchitecture}
\end{table}
\paragraph{Architecture for 2D shape experiments.} Table \ref{2dcnnarchitecture}
shows the CNN architecture used as the encoder. The input $I$ is an image of
size $64\times64$ and output $\Phi(I)$ is a vector of size $2048$.
Table~\ref{2d-rnnarchitecture} describes the architecture used in the decoder.
The RNN decoder is based on a GRU unit that at every time step takes as input
the encoded feature vector and previous instruction encoded as a $128$
dimensional vector obtained by a linear mapping of the $401$ dimensional one-hot
vector representation. At first time step, the previous instruction vector
represents the \texttt{START} symbol. Embedded vector of previous instruction is
concantenated with $\Phi(I)$ and is input to the GRU. The hidden state of GRU is
passed through two dense layer to give a vector of dimension $400$, which after
\texttt{softmax} layer gives a probability distribution over instructions. The
output distribution is over $396$ different shape primitives, $3$ operations
(\texttt{intersect}, \texttt{union} and \texttt{subtract}) and a \texttt{STOP}.
We exclude the \texttt{START} symbol from the output probability distribution.
Note that the circle, triangle or square at a particular position in the image
and of a particular size represents an unique primitive. For example, $c(32, 32,
16)$, $c(32, 28, 16)$, $s(12, 32, 16)$ are different shape primitives.
\paragraph{Architecture for 3D shape experiments.}
\begin{table}[]
  \centering
  \setlength{\tabcolsep}{1pt}
\begin{tabular}{l|c}
  \textbf{Layers}                                    & \textbf{Output}       \\ \hline
  Input Voxel                                           & $64$ $\times$ $64$ $\times$  $64$ $\times$ $1$  \\
  Relu(Conv3d: $4$ $\times$ $4$ $\times$ $4$, $1$ $\rightarrow$ $32$)           & $64$ $\times$ $64$ $\times$ $64$ $\times$ $32$  \\ 
  BN(Drop(Max-pool($2$ $\times$ $2$ $\times$ $2$)))                   & $32$ $\times$ $32$ $\times$ $32$ $\times$ $32$  \\ 
  Relu(Conv3d: $4$ $\times$ $4$, $32$ $\rightarrow$ $64$)                & $32$ $\times$ $32$ $\times$ $32$ $\times$ $64$  \\ 
  BN(Drop(Max-pool($2$ $\times$ $2$ $\times$ $2$)))                   & $16$ $\times$ $16$ $\times$ $16$ $\times$ $64$  \\ 
  Relu(Conv3d: $3$ $\times$ $3$, $64$ $\rightarrow$ $128$))              & $16$ $\times$ $16$ $\times$ $16$ $\times$ $128$ \\ 
  BN(Drop(Max-pool($2$ $\times$ $2$ $\times$ $2$)))                   & $8$ $\times$ $8$ $\times$ $8$ $\times$ $128$    \\ 
  Relu(Conv3d: $3$ $\times$ $3$, $128$ $\rightarrow$ $256$))             & $8$ $\times$ $8$ $\times$ $8$ $\times$ $256$    \\ 
  BN(Drop(Max-pool($2$ $\times$ $2$ $\times$ $2$)))                   & $4$ $\times$ $4$ $\times$ $4$ $\times$ $256$    \\ 
  Relu(Conv3d: $3$ $\times$ $3$, $256$ $\rightarrow$ $256$))             & $4$ $\times$ $4$ $\times$ $4$ $\times$ $256$    \\ 
  BN(Drop(Max-pool($2$ $\times$ $2$ $\times$ $2$)))                   & $2$ $\times$ $2$ $\times$ $2$ $\times$ $256$    \\ 
  Flatten                                               & $2048$  
\end{tabular}
\vspace{2mm}
\caption{\textbf{Encoder architecture for 3D shape experiments.} Drop: dropout layer, BN: batch-normalization layer and Drop: dropout layer with 0.2 probability.}
\label{3d-cnnarchitecture}
\end{table}
\begin{table}[]
  \centering
  \setlength{\tabcolsep}{2pt}
\begin{tabular}{l|l|c}
    \textbf{Index} & \textbf{Layers} & \textbf{Output}       \\ \hline
1 & Input shape encoding & $2048$ \\ 
2 & Input previous instruction & $6636$ \\ 
3 & Relu(FC($6636$ $\rightarrow$ $128$)) & $128$ \\ 
4 & Concatenate ($1$, $3$) & $2176$ \\ 
5 & Drop(GRU (hidden dim: $1500$)) & $1500$ \\ 
6 & Drop(Relu(FC($1500$ $\rightarrow$ $1500$))) & $1500$ \\ 
7 & Softmax(FC($1500$ $\rightarrow$ $6635$)) & $6635$
\end{tabular}
\vspace{2mm}
\caption{\textbf{Decoder network architecture for 3D shapes experiments.} FC: Fully connected dense layer, Drop: dropout layer with 0.2 probability. Dropout on
  GRU are applied on outputs but not on recurrent connections.}
\label{3d-rnnarchitecture}
\end{table}
Input to 3D shape encoder (3DCNN) is a voxel grid $I$ of size $64$ x $64$ x $64$
and outputs an encoded vector $\Phi(I)$ of size $2048$, as shown in the Table
\ref{3d-cnnarchitecture}. Similar to the 2D case, at every time step, GRU takes
as input the encoded feature vector and previous ground truth instruction. The
previous ground truth instruction is a $6636$-dimensional (also includes the
\texttt{start} symbol) one-hot vector, which gets converted to a fixed
$128$-dimensional vector using a learned embedding layer. At first time step the
last instruction vector represents the \texttt{START} symbol. Embedded vector of
previous instruction is concatenated with $\Phi(I)$ and is input to the GRU. The
hidden state of GRU is passed through two dense layers to give a vector of
dimension $6635$, which after Softmax layer gives a probability distribution
over instructions. The output distribution is over $6631$ different shape
primitives, $3$ operations (\texttt{intersect}, \texttt{union} and
\texttt{subtract}) and a \texttt{STOP}. We exclude the \texttt{START} symbol
from the output probability distribution. Similar to 2D case, $cu(32, 32, 16,
16)$, $cu(32, 28, 16, 12)$, $sp(12, 32, 16, 28)$ are different shape primitives.
Table \ref{3d-rnnarchitecture} shows details of decoder.

\subsection{Qualitative Evaluation}
\begin{figure*}[h]
\centering
\includegraphics[scale=0.4]{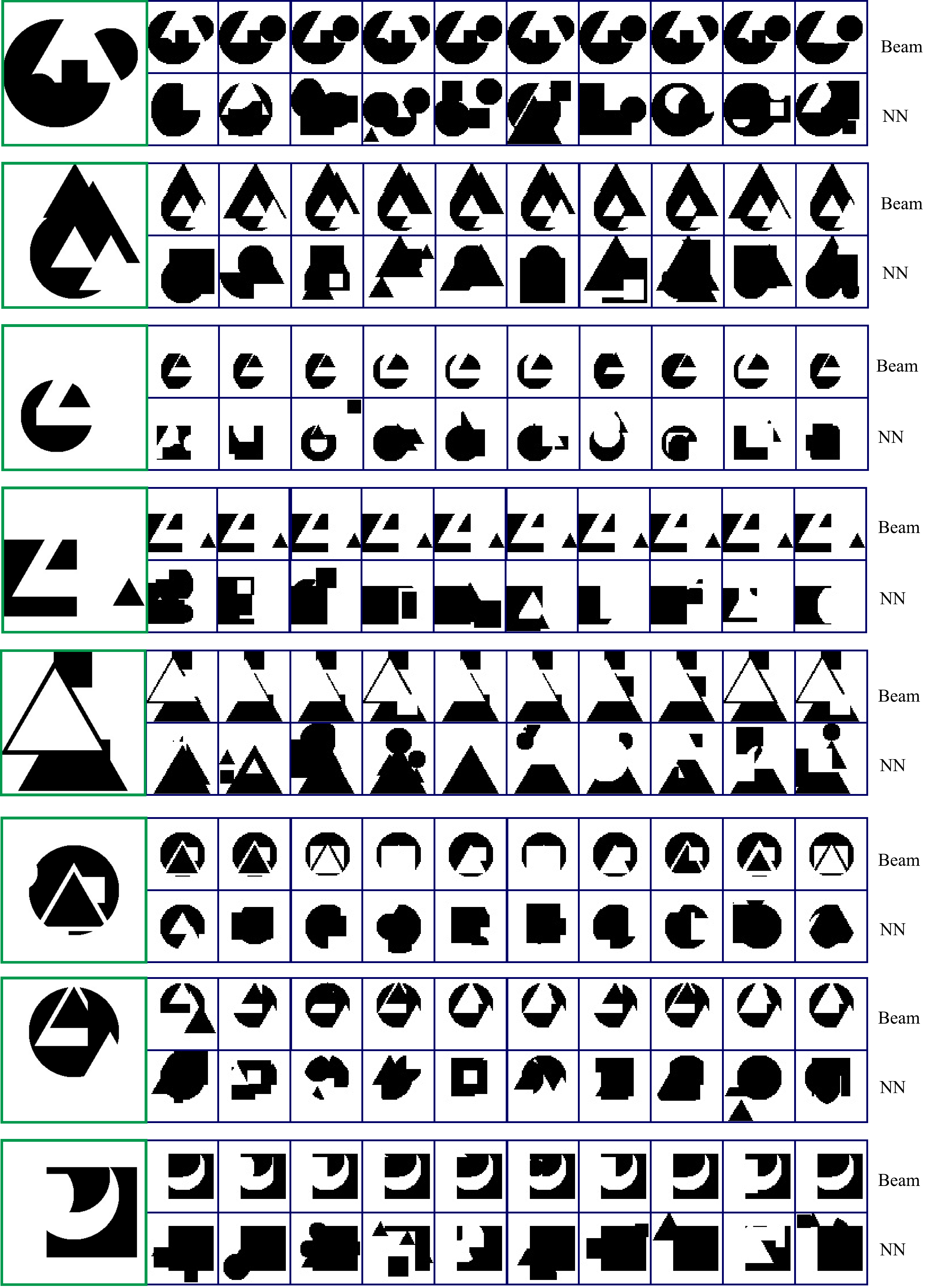}
\caption{\textbf{Qualitative evaluation on 2D synthetic dataset.} In green outline is the
groundtruth, top row represent top-10 beam search results, bottom row represents
top-10 nearest neighbors.}
\label{fig:synth2d-1}
\end{figure*}

\begin{figure*}[]
\centering
\includegraphics[scale=0.4]{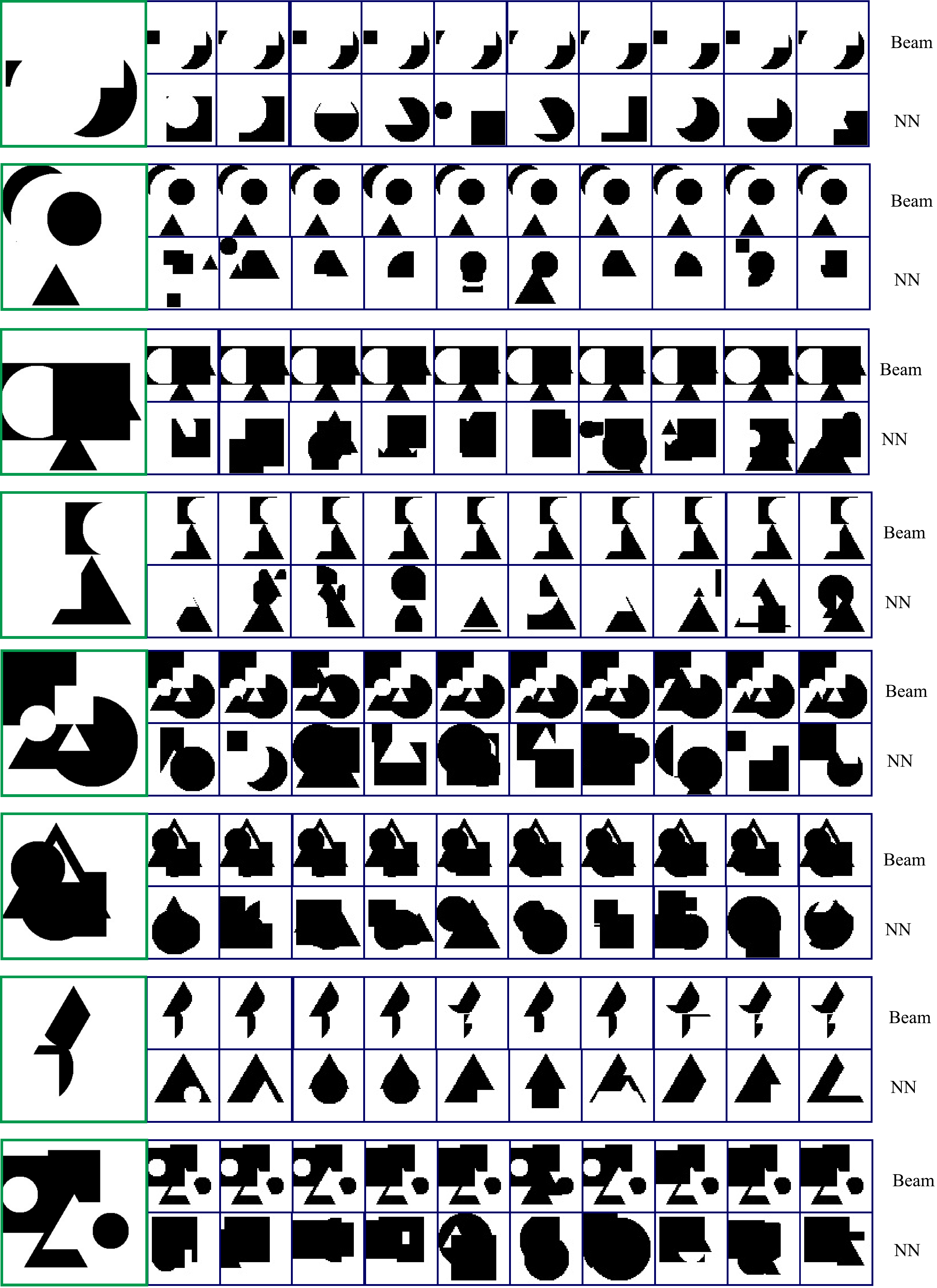}
\caption{\textbf{Qualitative evaluation on 2D synthetic dataset.} In green outline is the
groundtruth, top row represent top-10 beam search results, bottom row represents
top-10 nearest neighbors.}
\label{fig:synth2d-2}
\end{figure*}

\begin{figure*}[]
\centering
\includegraphics[width=\textheight, angle=90]{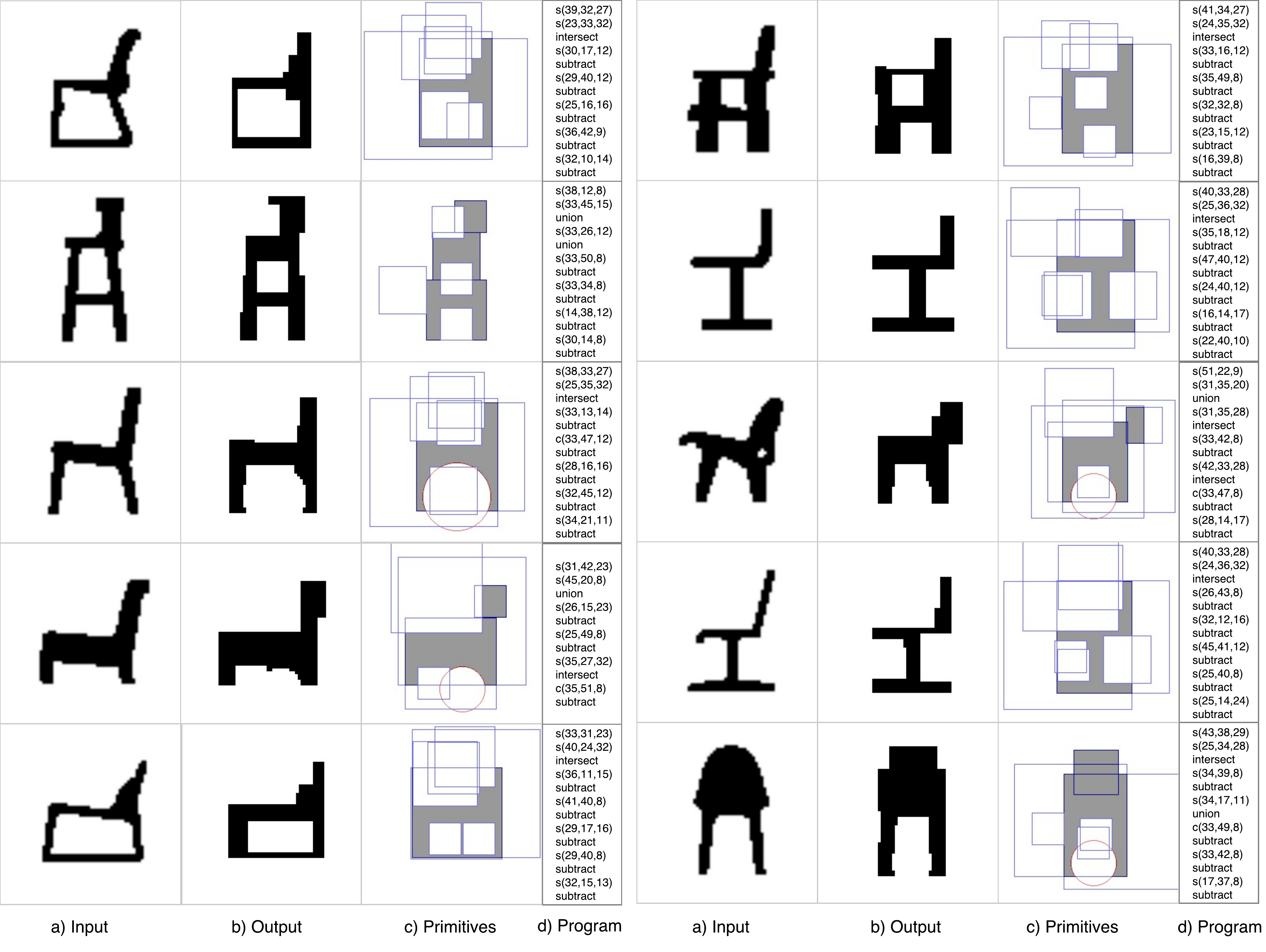}
\caption{\textbf{Performance of our full model on 2D CAD images}. a) Input image, b)
  output from our full model, c) Outlines of primitives present in the generated
  program, triangles are in green, squares are in blue and circles are in red d) Predicted program. $s$, $c$ and $t$ are shape primitives that represents $square$, $circle$ and $triangle$ respectively, and $union$, $intersect$ and $subtract$ are boolean operations.}
\label{fig:CAD1}
\end{figure*}

\begin{figure*}[]
	\centering
	\includegraphics[width=\textheight, angle=90]{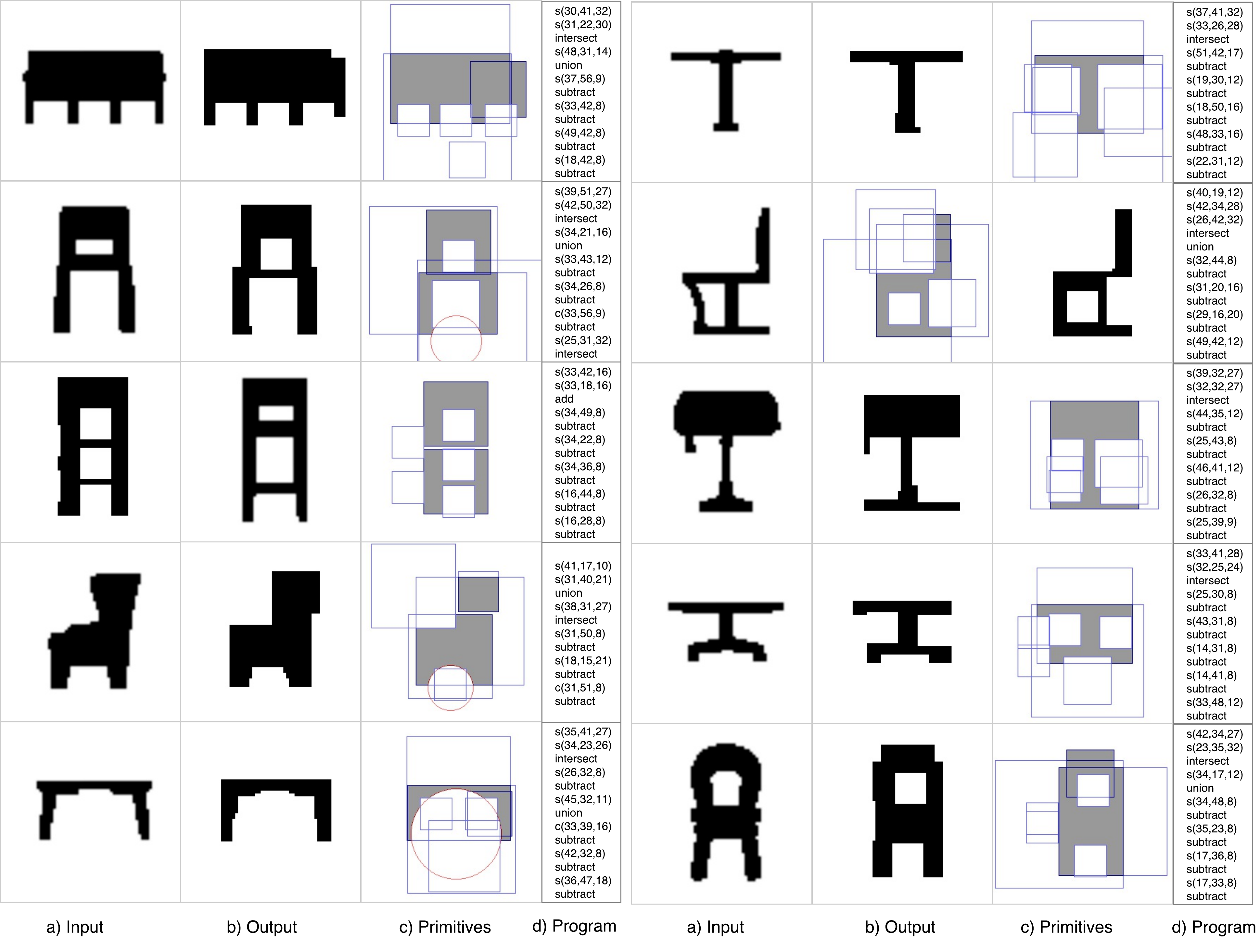}
\caption{\textbf{Performance of our full model on 2D CAD images}. a) Input image, b)
	output from our full model, c) Outlines of primitives present in the generated
	program, triangles are in green, squares are in blue and circles are in red d) Predicted program. $s$, $c$ and $t$ are shape primitives that represents $square$, $circle$ and $triangle$ respectively, and $union$, $intersect$ and $subtract$ are boolean operations.}
	\label{fig:CAD2}
\end{figure*}

In this section, we show more qualitative results on different dataset. We first
show peformance of our CSGNet trained using only Supervised learning on 2D
synthetic dataset, and we compare top-10 results from nearest neighbors and and
top-10 results from beam search, refer to the Figure \ref{fig:synth2d-1} and
\ref{fig:synth2d-2}. Then we show performance of our full model (using RL + beam
search + visually guided search) on CAD 2D shape dataset, refer to the
Figure~\ref{fig:CAD1} and ~\ref{fig:CAD2}.


\end{document}